\documentclass[10pt,twocolumn,letterpaper]{article}

\usepackage[pagenumbers]{cvpr} % To force page numbers, e.g. for an arXiv version

\usepackage{graphicx}
\usepackage{amsmath}
\usepackage{amssymb}
\usepackage{booktabs}
\usepackage{pifont}
\usepackage{xcolor}

\usepackage[pagebackref,breaklinks,colorlinks]{hyperref}

\usepackage[capitalize]{cleveref}
\crefname{section}{Sec.}{Secs.}
\Crefname{section}{Section}{Sections}
\Crefname{table}{Table}{Tables}
\crefname{table}{Tab.}{Tabs.}

\newcommand{\myparagraph}[1]{\vspace{6pt}\textbf{#1}}
\newcommand{\detector}{\mathcal{F}}
\newcommand{\known}{\mathcal{Y}}
\newcommand{\unknown}{\mathtt{u}}
\newcommand{\approach}{UNKAD} % after command add \ to add a space 
\newcommand{\neuron}{Direct Prediction}
\newcommand{\real}{{\rm I\!R}}

\begin{document}

\title{\vspace{5pt}Detecting the unknown in Object Detection}

\author{
{\vspace{2pt}}
Dario Fontanel$^{1}$, Matteo Tarantino$^{1}$, Fabio Cermelli$^{1,2}$, Barbara Caputo$^{1}$\\
$^1$Politecnico di Torino, $^2$Italian Institute of Technology\\
{\tt\small dario.fontanel@polito.it} \\
{\vspace{2pt}}
}

\maketitle

%%%%%%%%% ABSTRACT
\begin{abstract}
Object detection methods have witnessed impressive improvements in the last years thanks to the design of novel neural network architectures and the availability of large scale datasets.
However, current methods have a significant limitation: they are able to detect only the classes observed during training time, that are only a subset of all the classes that a detector may encounter in the real world. Furthermore, the presence of unknown classes is often not considered at training time, resulting in methods not even able to detect that an unknown object is present in the image.
In this work, we address the problem of detecting unknown objects, known as open-set object detection. We propose a novel training strategy, called UNKAD, able to predict unknown objects without requiring any annotation of them, exploiting non annotated objects that are already present in the background of training images. In particular, exploiting the four-steps training strategy of Faster R-CNN, UNKAD first identifies and pseudo-labels unknown objects and then uses the psuedo-annotation to train an additional unknown class.
While UNKAD can directly detect unknown objects, we further combine it with previous unknown detection techniques, showing that it improves their performance at no costs.
\end{abstract}

%%%%%%%%% BODY TEXT
\section{Introduction}
\label{sec:introduction}
\begin{figure}[tb]
  \centering
  \includegraphics[width=\columnwidth]{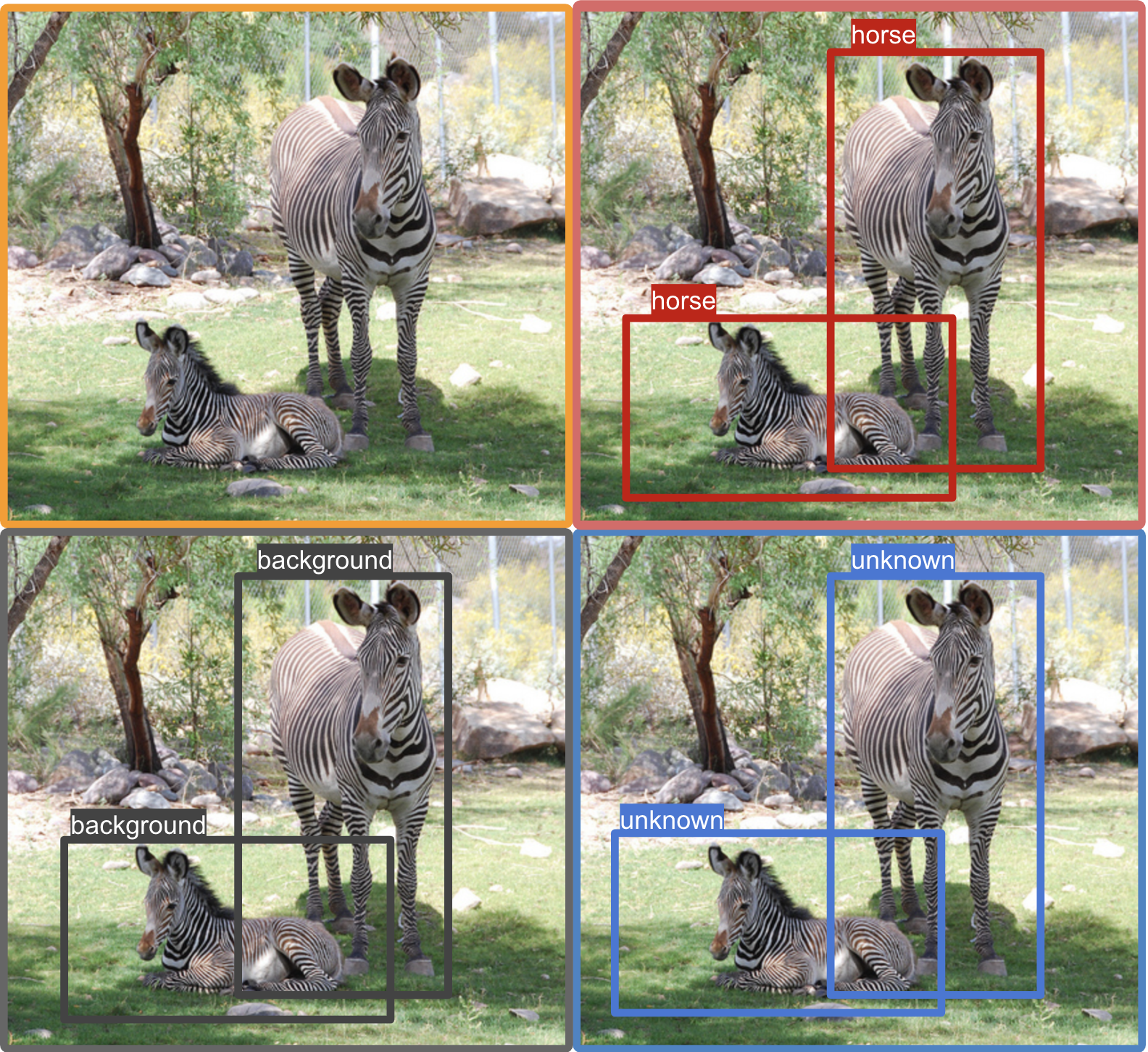} 
%   \vspace{-20pt}
   \caption{Open-set detection aims to detect previously unseen objects. While previous works consider the unknown objects as background, we are interested in evaluation the capability of detectors to predict unseen object as \textit{unknown}. From left to right, we can see the in the first row a MS-COCO \cite{lin2014microsoft} test image and the wrong prediction of Faster R-CNN \cite{ren2015faster}. In the second one, we can see the background prediction of \cite{dhamija2020overlooked}, and the unknown detection made by our framework.}
%   \vspace{-20pt}
  \label{fig:teaser}
\end{figure}

For autonomous system acting in the real world, it is essential to have a clear understanding of their surroundings. To accomplish this purpose, multiple works have concentrated on the task of object detection \cite{lin2017feature, girshick2014rich,girshick2015fast,he2017mask,ren2015faster, wang2019learning, liu2016ssd, lin2017focal, redmon2017yolo9000, redmon2016you, cao2019hierarchical}, where the goal is to locate the objects inside an image and to assign them a category. 
Despite the outstanding performance demonstrated by state-of-the-art detection models \cite{ren2015faster, he2017mask, liu2016ssd, redmon2016you}, they still have a critical limitation: they are able to predict only the classes that they have observed during training time, which are defined a priori and annotated in the training dataset. Regardless of how extensive the training dataset is, it is practically impossible to capture all the possible objects that system might ever encounter. Thus, in the real world, when current detection models are presented with an \textit{unkown} object they will not detect it, considering it either as a known class or as background.
For example, let us consider an autonomous driving car that has been trained on classes that are likely to be encountered in the city, such as pedestrians, other vehicles, trees, etc. However, at some point in time, the car is lead to a country region and a wild animal appears on the road. If we do not enable detection models to recognize unknown objects, the detectors would not be able to spot the unknown animal and will label it as background, putting at risk the safety of the passengers since the car cannot recognize the obstacle and may crash on it. 
%However, an unavoidable limitation for object detection models is the amount of annotated classes at training time. Regardless of how extensive the training dataset is, it is practically impossible to capture all the possible objects that system might ever encounter. 
Ideally, we would like to have object detection models able to detect all the objects in the world ad possibly understand whether an object corresponds to one of the training categories or if it is an unknown object that they have never seen before. 

In this work, we focus on this task, known in the literature as open-set object detection, for which an illustration is depicted in Figure~\ref{fig:teaser}. % The goal of open set detection is to recognize unknown objects, being unseen during the training phase of the model.
Previous works \cite{dhamija2020overlooked} addressed the problem partially, focusing only on limiting performance degradation on known classes when unknown data is encountered. We believe, as demonstrated in the previous example, that considering unknown objects as background is not sufficient to produce systems able to act in the real world as they could not detect obstacles,  introducing safety risks.
Differently, we propose to explicitly detect unknown objects at test time using a novel approach called \approach. It exploits the fact that, in object detection, multiple objects are preset in an image but only few of them are annotated, since the others are not considered relevant. %to learn unknown category at training time. 
By exploiting the multi-steps training strategy of the Faster R-CNN \cite{ren2015faster}, \approach\ is able to extract pseudo-supervision for the unknown objects, identifying them in the background areas of the training images. Moreover, it exploits the pseudo-supervision on the classification head, introducing an additional \textit{unknown} class that can be predicted, which helps to learn sharper decision boundaries between known and unknown objects. 
While this training strategy is already able to detect unknown objects directly predicting them in the classifier, we combine it with standard out-of-distribution methods and we show that it is able to improve the performance of them with respect to a standard Faster R-CNN training strategy.
%allows to introduce a simple yet effective unknown detection strategy which directly predicts the unknown from the model itself. 
We demonstrate the performance of \approach\ on the popular Pascal VOC 2007 \cite{Everingham2009ThePV} and MS-COCO \cite{lin2014microsoft} benchmarks.

\myparagraph{Contributions.} To summarize, our contributions are:
\begin{itemize}
    \item We propose a novel perspective for open-set detection problem, developing a simple yet effective training strategy;
    \item We introduce out-of-distribution standard approaches into the object detection framework, leading to novel analysis with respect to previous benchmarks;
    \item Experiments on the widely adopted Pascal VOC and MS-COCO datasets show that standard out-of-distribution methods benefits from our approach by a significant amount.
\end{itemize}

\section{Related work}
\label{sec:related}
In this section, we review the foundations of our work, \ie object detection, open-set and open-set object detection.

\myparagraph{Object detection.}
Modern object detection approaches \cite{lin2017feature, girshick2014rich,girshick2015fast,he2017mask,ren2015faster, wang2019learning, liu2016ssd, lin2017focal, redmon2017yolo9000, redmon2016you, cao2019hierarchical} are dominated by architectures based on convolutional neural networks that differ on whether or not candidate object proposals are used.  
We can group these works in two different categories: two-stage approaches \cite{lin2017feature,girshick2014rich,girshick2015fast,he2017mask,ren2015faster}, that generates object proposals which are classified and regressed by a region-of-interest (RoI) head module, and single-stage approaches \cite{wang2019learning, liu2016ssd, lin2017focal, redmon2017yolo9000, redmon2016you, cao2019hierarchical} that simultaneously output both classification scores and regressed bounding boxes without the need of any object proposal. Despite the outstanding performance achieved on popular benchmarks \cite{Everingham2009ThePV, lin2014microsoft}, all of these architectures operate solely in an offline fashion, which means that after the model has been trained, additional knowledge cannot be added. Despite recent efforts to advance and deal with the inclusion of novel classes \cite{shmelkov2017incremental, peng2020faster, joseph2021towards}, none of these techniques strictly focus on open-set object detection.

\myparagraph{Open-set.}
In recent years, open-set recognition has sparked a lot of interest in the machine learning community \cite{bendale2015towards,perera2020generative,zhou2021learning}. The seminar work of \cite{bendale2015towards} formalized the open-set recognition task
% introducing a variation of SVM that sculpts the decision space to better discriminate between samples that belong to the training distribution (also known as \textit{in-distribution} or \textit{known} data) and samples that do not (also noted as \textit{out-of-distribution} or \textit{unknown} data).
investigating for the first time what might happen when a model is ask to recognize data from categories that it has never seen before. One of the most popular open-set sub-field is out-of-distribution (OOD) detection \cite{hendrycks2016baseline,liang2018enhancing,hsu2020generalized} which aims at discriminating between samples that belong to the training distribution (also known as \textit{in-distribution} or \textit{known} data) and samples that do not (also noted as \textit{out-of-distribution} or \textit{unknown} data).

\cite{hendrycks2016baseline} settled the standard baseline for out-of-distribution detection by applying a threshold over the maximum softmax probability (MSP) to categorize a sample as belonging to known classes or as an unknown one. \cite{gal2016dropout, kendall2017uncertainties} used Monte Carlo Dropout (MC-Dropout) to estimate the model uncertainty by forwarding the same image through the network multiple times, each time with a different dropout probability. \cite{corbiere2019addressing} trains an additional neural network to output high confidence values when the prediction of the main model on in-distribution data is correct. This additional branch is then used to detect if the network prediction is reliable or not. Scaling the softmax probabilities by a temperature, for each sample \cite{liu2020energy} computes the energy which is higher for known samples rather than unobserved ones. ODIN \cite{liang2018enhancing} further enhanced MSP by introducing a temperature scaling factor in the softmax function and small perturbations over the test images. Both these hyperparameters are learned on an OOD validation set available during training. As collecting OOD data is not always feasible, in this work we avoid relying on it, leveraging background objects to model unknown properties.

\myparagraph{Open-set object detection.} 
The open-set framework introduces additional challenges once adopted to the object detection task. \cite{miller2018dropout} has been the first to bring open-set object detection to light and \cite{dhamija2020overlooked} further investigated the problem, assessing how detectors performance on known classes varies when evaluated on both known and unknown objects. In this paper, instead, we believe it is critical to evaluate also the capability of object detection models of recognizing objects as unknown. As a result, we need to introduce in the object detection framework out-of-distribution approaches able to distinguish between known and unknown categories, evaluate models' performance on both. Recently, few works \cite{gupta2021ow, joseph2021towards} made a step further introducing into object detection models the rejection capability, \ie the ability of recognizing an object as unknown. To detect unknown samples, \cite{joseph2021towards} employed a contrastive approach while \cite{gupta2021ow} adopted a transformer-based architecture. Despite the introduction of the rejection capability, it is worth noting that in this work we do not provide comparisons with \cite{joseph2021towards} and \cite{gupta2021ow} as their primarily concerned is with building models capable of expanding their knowledge over time, hence de facto focusing on another task and objective.

\section{Method}
In this section we first formalize the problem definition and the importance of distinguishing between known e unknown categories (Section ~\ref{sec:problem}), We then describe \approach, showing how to detect unknown objects during training and how to involve them through the learning process (Section ~\ref{sec:pseudo}). Finally, in Section ~\ref{sec:rejection} we will analyze and compare different rejection strategies. 
%An illustration of \ours\ is provided in Fig.~\ref{fig:method}.

\subsection{Preliminaries and problem formulation} \label{sec:problem}
The goal of open-set object detection is to detect objects that have not been seen during the training phase \cite{dhamija2020overlooked}. To perform this task, the model is provided with a training dataset $\mathcal{T}_{train}=\{(x, y)\}$, where $x$ is an image and $y$ is its corresponding ground-truth label. As in standard object detection, $y$ contains bounding box annotations for a set of classes, that we indicate as \textit{known} classes $\known$. We note that, in object detection, the training images contains multiple objects, but not all of them are annotated. We denote all the objects not annotated as \textit{unknown} objects. 
% \comment{We also define as $\bkg$ the set of objects that labelled as background in the training set as they are not explicitly meant to be learned as classes of interest, \ie $\mathcal{K}_{train} = \known \cup \bkg$}. 
During testing, we are provided with a dataset $\mathcal{T}_{test}$ containing objects of both $\known$ and never seen categories (\textit{unknown}). 
%, \ie $\mathcal{K}_{test} = \known \cup \unknown$. 

Focusing on R-CNN architectures \cite{girshick2015fast, ren2015faster}, our aim is to learn a function $\detector$ mapping an image $x$ to its known and unknown predictions at bounding box level. We consider $\detector$ as built upon four key components. The first one is the feature extractor $\detector_{FE}$ that is responsible for producing a feature map for each image $x$.
% \in \real^{K \times H \times W}$, indicating respectively the number of channels $K$, the height of the image $H$ and its width $W$. 
The second one is a region proposal network $\detector_{RPN}$ that is fed with the feature map of the input image and produces a set of rectangular object proposals associated with a binary objectness score. In particular, it firstly projects the input feature map into a lower dimensional space by means of an intermediate projection layer and then the projected features are fed into two separate convolutional layers, one responsible for regressing the bounding box and the other responsible to output an objectness score $\omega$. The set of object proposals is then applied to the feature map and pooled, producing $\mathcal{R}$ regions of interest (RoIs). The third component is the classification head and aims to classify the objects and regress the correct bounding box. It is composed by two functions: a class-scoring function 
% The third one is a scoring function
$\mathcal{F}_{\psi}:\mathcal{R}\rightarrow\real^{\mathcal{R} \times |\known|}$ that maps $\mathcal{R}$ RoIs to a box-level class scores, and a function $\mathcal{F}_{\rho}$ that regresses, for each RoIs and class, the precise bounding box coordinates, \ie $\mathcal{F}_{\rho}:\mathcal{R}\rightarrow\real^{\mathcal{R} \times 4|\known|}$.
% The final component is the known detection function $\mathcal{F}_{\omega}:\real^{|\mathcal{R}|\times|\mathcal{K}_{known}|}\rightarrow \real^{|\mathcal{R}|}$ that maps class scores to class probabilities; while 
Finally, the last component is the unknown detection function $\mathcal{F}_{\phi}:[0,1]^{\mathcal{R}\times|\mathcal{K}_{known}|}\rightarrow \real^{\mathcal{R}}$ that a binary score indicating if the RoI is unknown ($\mathcal{F}_{\phi}=1$) or not ($\mathcal{F}_{\phi}=0$). 

\subsection{Detecting the unknown} \label{sec:detect}
% Many out-of-distribution detection approaches \cite{lee2018simple,hendrycks2018deep,vyas2018out,dhamija2018reducing,li2020background, mohseni2020self,yu2019unsupervised,lee2018simple} leverage external OOD datasets as extra supervision in order to learn richer decision boundaries between $\known$ and $\unknown$ samples. Clearly, collecting external unknown datasets is not always feasible, and most of the time human intervention is required to at least partially annotate them. Hence, in this paper we take a different direction. Avoiding the need of OOD data, we present an approach able to learn unknown class objects during training by exploiting pseudo-labelling and $\detector_{RPN}$ properties. We named this approach \textit{\approach}.
% We start from the intuition that $\detector_{RPN}$ is a class-agnostic detector first and foremost able to establish whether an object is present within a certain region. This behaviour is experimentally confirmed in both closed- and open-set conditions (see Section~\ref{sec:ablation}), where $\detector_{RPN}$ is asked to separate the generic background environment from $U_{K}$ known-unknown objects in the former and from $U_{U}$ unknown-unknown in the latter. To benefit from this property, we need a threshold to establish us whether a detection is an actual object or not. To accomplish this, after the first and the third steps of the Faster R-CNN training procedure, we derive the objectness threshold $\tau_{obj}$ as:
In this work we take inspiration from out-of-distribution detection approaches \cite{lee2018simple,hendrycks2018deep,vyas2018out,dhamija2018reducing,li2020background} that leverage external OOD data as extra supervision to learn richer decision boundaries between known and unknown samples. As collecting external unknown datasets is not always feasible, and most of the time human intervention is required to at least partially annotate them, our intuition is that we can leverage as OOD data a portion of the objects labelled as background. This is uniquely possible due to the object detection framework, that provides labels only for the classes of interest in $\known$, leaving all the other objects that are not meant to be learned not annotated. 

\begin{figure}[tb]
  \centering
  \includegraphics[width=\columnwidth]{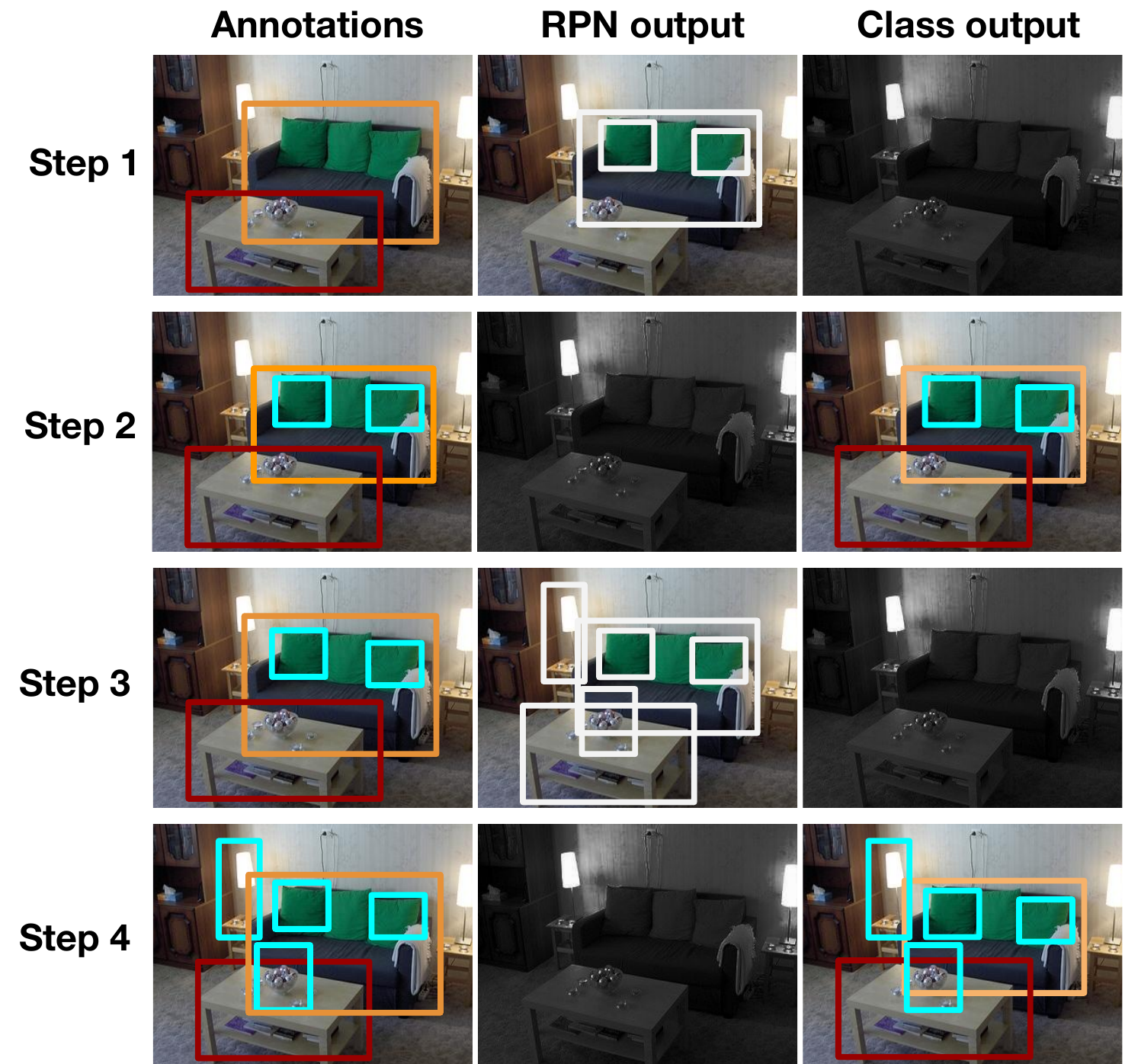} 
%   \vspace{-20pt}
   \caption{Illustration of the \approach\ training procedure. In the first step, the RPN learns to predict class-agnostic RoIs (white boxes) containing objects. It then pseudo-labels unknown objects and, in step 2, trains the classifier with the additional unknown class. In step 3, the RPN is trained again, considering both known and unknown objects. Finally, in step 4 the final model able to detect both known and unknown objects is obtained.}
%   \vspace{-20pt}
  \label{fig:method}
\end{figure}

Towards this end, we train our $\detector$ detector module in a four alternate steps strategy, adapting the methodology proposed in Faster R-CNN \cite{ren2015faster} and proposing \textbf{\approach\ }(\textbf{UNK}nown \textbf{A}ware \textbf{D}etection), that extends the model classification ability towards unknown objects. In particular, in the first and the third steps the $\detector_{RPN}$ learns to extracts class-agnostic RoIs, while in the second and the fourth the detector $\detector$ learns to classify known classes $\known$ and the unknown. %In the following, we detail how we couple this alternate strategy with the unknown learning objective at training time. 
An illustration of \approach\ four steps training strategy is reported in \cref{fig:method}.
% We invite the reader to refer to \cite{girshick2015fast, ren2015faster} for additional details on Faster R-CNN training and architecture. 
% We named this approach ).

\myparagraph{Discovering the unknown.} 
The intuition behind \approach\ is that unknown objects are already present in the training images and we can learn to recognize them by using the RPN class-agnostic ability to detect objects. %pseudo-annotating background training objects. 
%We start from the experimental analysis (see Section~\ref{sec:ablation}) confirming that the class-agnostic $\detector_{RPN}$ is able to separate the generic background from unknown objects that are contained into it. 
In particular, the RPN exploits the ground truth labels to find and predict the RoI, \ie n region where is likely to find an object. During preliminary experiments (see Section~\ref{sec:ablation}), we found that the RPN was able to detect, other than the annotated objects, the objects in the image background with high confidence. Arguing that its class-agnostic nature helps to find any object in the image, we propose to pseudo-label the objects detected in the background as \textit{unknown} objects.
%To actually learn them, however, we need to extract pseudo-annotations, that can be used to train the classification head.
In order to generate pseudo-labels during training, we propose to use a simple thresholding mechanism. In particular, we define a threshold to establish whether a RoI is a region containing an object (known or unknown) or not. 
The threshold $\tau_{obj}$ is derived from data, and it is computed as:
\begin{equation}
        \tau_{obj} = \mu + \lambda \cdot \sigma,
\end{equation}
where
\begin{equation}
    {\mu = \frac{\sum_{i}{\omega_i}}{|\mathcal{R}_{FG}|}},
\end{equation} 
\begin{equation}
    {\sigma = \sqrt{\frac{\sum_{i}{(\omega_i - \mu)^2}}{|\mathcal{R}_{FG}|} }},
\end{equation} 
and $\mu$ and $\sigma$ represent respectively the mean and the standard deviation of foreground-layer activations, $\mathcal{R}_{FG}$ is the set of RoIs that either (i) has the highest IoU, or (ii) has an IoU overlap with with any ground-truth box higher than $0.7$. We recall that $\omega_i$ represents the objectness score for the $i^{th}$ region fed to $\detector_{RPN}$. $\lambda$ is hyperparameter set to 1. 

%To accomplish this, after the first and the third steps
%we consider all the RoI with an objctness score greater than ${\tau}_{obj}$ as actual objects.
% we are able to learn an \textit{unknown} object class directly during training. 
% Furthermore, we label as unknown class $\unknown$ all the RoIs that have i) an IoU lower than $0.3$ with all the ground-truth boxes and ii) foreground-layer $w_{cls}$ activations higher than ${\tau}_{obj}$.
To obtain the pseudo-labels, we first select all the RoIs having an objectness score greater than ${\tau}_{obj}$. Then, we select all the RoIs that do not match with a ground truth annotation (\ie they do not  overlap more than 0.3 IoU with it) ad we associate them to the unknown class $\unknown$. 

Once we obtain pseudo-labels on the unknown objects that are present in the training images, we exploit them to train the classification head to detect every unknown object. 

\myparagraph{Learning to predict unknown} \label{sec:pseudo}
As our primary goal is to learn an unknown detector through pseudo-supervision, we add to the final classification layer $\mathcal{F}_{\phi}$ an additional class that leverages the unknown pseudo-labels generated by $\detector_{RPN}$. More precisely, following the alternate training strategy of Faster R-CNN \cite{ren2015faster, girshick2015fast}, at the end of the first and the third steps, we perform the pseudo-labelling phase. In the second and the fourth steps, $\detector_{\rho}$ and $\detector_{\phi}$ are learned, leveraging both annotated and pseudo-annotated data. By using the pseudo-annotations as ground truth, $\detector$ learns to separate at training time the unknown category from all the known classes. %\dario{aggiungere cross entropy?}

\subsection{Rejection strategies} \label{sec:rejection}
%\comment{At test time usiamo la standard pipeline di faster rpn->roi ->classificatore e anomaly score. Se l'anomaly score è sopra threshold la consideriamo anomalia, altrimenti usiamo l'output del classificatore per predirre una class o bkg
%dario{fare exp: direct prediction e mettere threshold=max prob classes esclusa unk class}}
In this section we present different strategies to predict a RoI as an unknown object, \ie different implementation of the $\detector_{\phi}$ function. In particular, we assume that, at inference time, we obtain $\mathcal{R}$ RoIs from the RPN, which are passed to the classification head to obtain the final classification scores $s$ using $\detector_{\psi}$. 
% detection we discuss how standard approaches differ in computing known and unknown probabilities with $\detector_{\phi}$ function.

\myparagraph{Direct prediction.}
The standard approach of localizing and classifying objects in closed-set scenarios is to localize the object and assign it the class with the highest class probability.
Since \approach\ extends the classification logits also to the unknown class, we may apply the same principle and predict a sample as unknown only if has the unknown class has the highest score.
% However, in open-set evaluations, the argmax operation is not directly applicable without any additional computation, as it is not possible to directly predict the unknown class from the model itself. With our \approach\ approach we are able to directly produce a score also for the unknown class. 
Formally, given an image and a RoI $r$, we obtain the set $s$ of class scores, including also the score for the $\unknown$ class, and we compute $\detector_{\phi}$ as:
\begin{equation} \label{sec:argmax}
    \detector_{\phi}(r) = 
    \begin{cases}
    1 \;\;& \text{if}\ \hat{y} = \unknown \\
    1 \;\;& \text{if}\ \hat{y} = \mathtt{b} \And \omega_r > \tau_{obj}  \\
    0 \;& otherwise
    \end{cases}
\end{equation}
where $\hat{y} = \operatorname*{arg\,max}_{c \in \known \cup \unknown} s_c$ is the class with highest score, $\omega_r$ is the RPN objetness score for $r$, and $\mathtt{b}$ indicates the background class.
We note that we added the second case to avoid missing potential unknown objects. 

\myparagraph{Maximum Softmax Probability (MSP).}
Maximum Softmax Probability \cite{hendrycks2016baseline} leverages the highest probability value assigned to any known class in $\mathcal{Y}$ as a measure of uncertainty. % of the model on unknown data. 
For an image, given a RoI $r$ and the class scores $s$ for it, MSP computes the value for the unknown class $\detector_{\phi}(r)$ as:
\begin{equation}
    \label{sec:msp}
    \detector_{\phi}(r) = \max_{c\in\known\cup\mathtt{b}} \frac{e^{s_c}}{\sum_{k\in\known \cup \mathtt{b}} e^{s_k}} \;\; \leq \tau_{MSP},
\end{equation}
where $\tau_{MSP}$ is a user defined threshold that we set to $0.5$ and $\mathtt{b}$ indicates the background class. Intuitively, if a class or background is predicted with a probability inferior to $\tau_{MSP}$, MSP identifies the RoI as an unknown object.

% Despite its effectiveness, we argue that using softmax probabilities is not the best choice to estimate the anomaly scores. In fact, the softmax function may smooth the confidence of the model prediction on each pixel, leading to consider uncertain (and thus anomalous) pixels even with high predicted initial scores. 

\myparagraph{Energy-based classifier (Energy).}
The energy-based scoring function \cite{liu2020energy} adopts a completely different perspective from traditional classifiers. Instead of computing class probabilities, it maps each samples to a scalar value, \ie the energy.
Given an RoI $r$ of a image, $\detector_{\phi}(r)$ is computed as:
\begin{equation}
    \label{sec:energy}
    \detector_{\phi}(r) = - T \cdot \log \sum_{c=1}^{\known \cup \mathtt{b}} e^\frac{s_c}{T} \;\; \leq \tau_{EN},
\end{equation}
where T is the temperature hyperparameter, $\tau_{EN}$ is a user defined threshold set to $-3$, and $s_c$ is the classification score for class $c$. 
To align with the convention introduced by \cite{liu2020energy}, we keep the negative notation for the energy score, which is higher for known classes and lower for unknown ones.
 
\myparagraph{ODIN.} To enhance MSP performances, ODIN \cite{liang2018enhancing} adopted temperature scaling and input perturbation. In detail, before feeding the scores to the softmax function, ODIN scales each class scores by a temperature parameter $T$. In addition, it also pre-processes each input image $x$ by introducing a small perturbation, that we adapt in the context of object detection as:
\begin{equation}
    \label{sec:odin_perturb}
    \tilde{x} = x - \epsilon \cdot {sgn}( -\nabla_{x} \cdot \frac{1}{\mathcal{R}}\sum_{r=1}^{\mathcal{R}} \log(\max_{c\in\mathcal{Y}\cup\mathtt{b}} p_c)),
\end{equation}
where $\epsilon$ is the perturbation magnitude, $\nabla_{x}$ is the gradient vector with respect to $x$, and $p$ is the softmax of the scores $s$, \ie $p = \text{softmax}(s)$.
%$s_T$ is the maximum softmax probability computed using class scores scaled by temperature $T$.

As for MSP \cite{hendrycks2016baseline}, given a perturbed image $\tilde{x}$ and the class scores $s$ computed on the RoI $r$, ODIN computes $\detector_{\phi}(r)$ as:
\begin{equation}
    \label{sec:msp}
    \detector_{\phi}(r) = \max_{c\in\known} \frac{e^{s_c}}{\sum_{k\in\known} e^{s_k}} \leq \tau_{ODIN},
\end{equation}
$\tau_{ODIN}$ is a user defined threshold set to $0.4$.

\section{Experiments}
\label{sec:experiments}

% We conduct our experiments on ... which contains 5125 training images with paired semantic labels, 1031 validation images without anomalies and 1500 test images with anomalies. Test images contain anomalies randomly selected from a set of 250 objects. These objects are placed in the test images trying to reproduce plausible road scenarios. Figure \ref{fig:dataset} shows examples of images taken from the dataset.

% On this benchmark we compare our method with state-of-the-art anomaly segmentation approaches, namely MSP \cite{hendrycks2016baseline}, MSP + CRF \cite{hendrycks2019benchmark}, an auto-encoder (AE) based approach \cite{baur2018deep}, Dropout \cite{gal2016dropout}, and the generative approach SynthCP \cite{xia2020synthesize}. 

\begin{figure*}[t]
    \centering
    \includegraphics[width=0.99\linewidth]{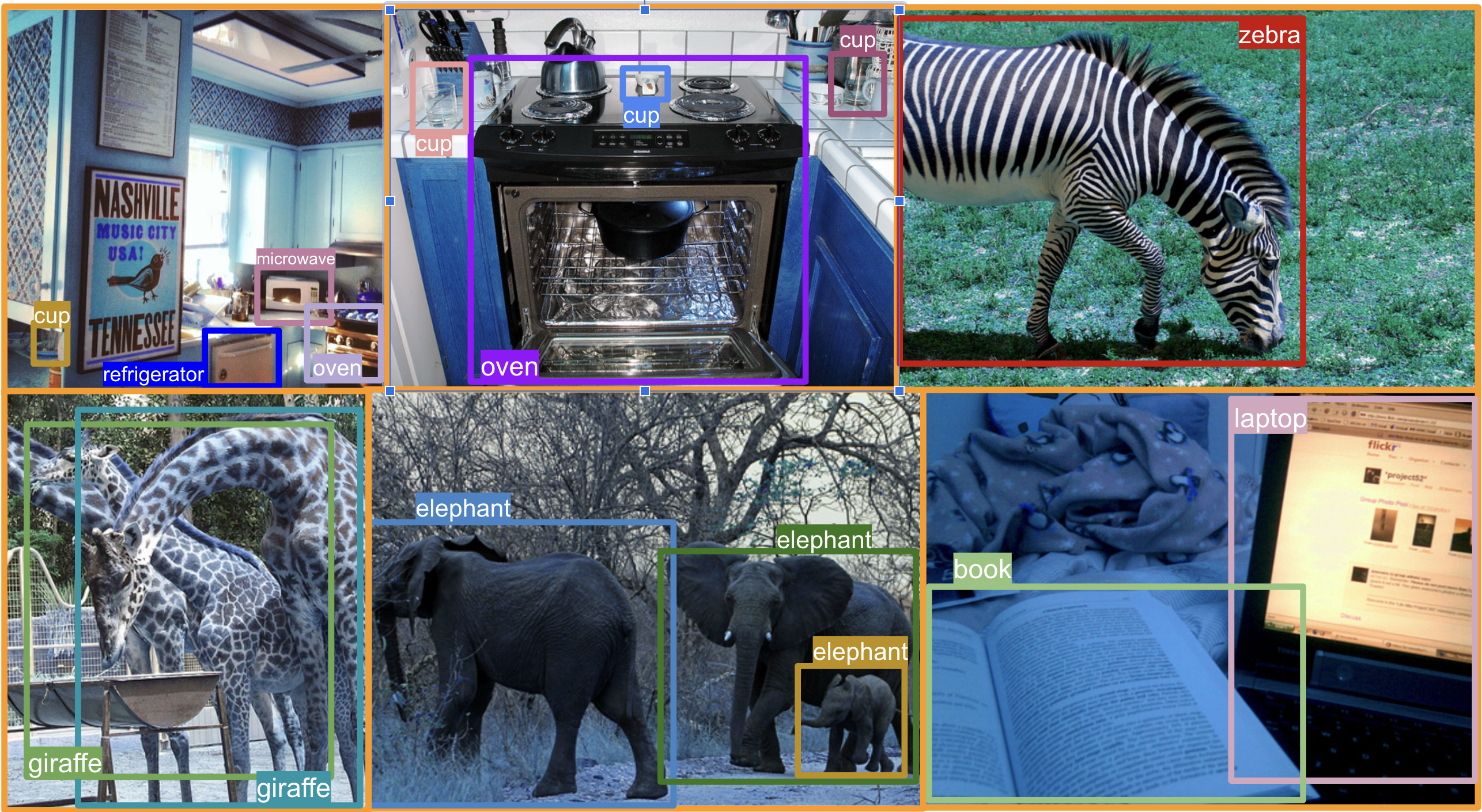}
    % \vspace{-5pt}
     \caption{Extract of unknown images taken from \cite{lin2014microsoft}.}
    \vspace{2pt}
    \label{fig:unknown}
 \end{figure*}\subsection{Experimental Protocol}

\myparagraph{Datasets.} 
We conduct our experiments on the popular Pascal VOC 2007 \cite{Everingham2009ThePV} and MS-COCO \cite{lin2014microsoft} datasets.  %following the evaluation split %\dario{non cito training, perché di fatto il training è diverso perché loro usano anche voc12. Solo il test split della WI è rispettato} 
Pascal VOC 2007 is a widely used benchmark that includes 20 foreground object classes and consists in 5K images for training, considering both training and the validation splits, and 5K for testing. 
MS-COCO is a large scale dataset that provide 80K images for training, 20K for validation and for the test. It contains annotation for 80 object classes of which 20 are in common with Pascal VOC 2007.
% The training set counts around 5k images, using both the training and the validation splits of Pascal VOC 2007.
For the open-set evaluation, we follow the split defined as WR$_{1}$ in \cite{dhamija2020overlooked}. 
In particular, the test set of Pascal VOC 2007 is used for the evaluation on the $\known$ classes, while 4952 images from MS-COCO training set that do not contain any Pascal VOC class are selected for the evaluation on the $\unknown$ class (see. Figure \ref{fig:unknown}), resulting in a total of nearly 10k test images.

\myparagraph{Baseline.}
We compare our proposed training strategy with the 4-steps Faster R-CNN \cite{ren2015faster} training procedure. We implement both strategies upon multiple open-set and out-of-distribution detection strategies. In particular, we implement the training strategy described in \cref{sec:rejection}: direct prediction, 
%On this benchmark we compare \approach with open set and out-of-distribution detection baselines, namely
MSP \cite{hendrycks2016baseline}, Energy \cite{liu2020energy} and ODIN\cite{liang2018enhancing}.
We note that it is not possible to use direct prediction on the standard Faster R-CNN training since it does not provide the unknown class in the classification head.

\myparagraph{Architectures.}
% We report experiments for each rejection strategy built upon both the standard training objective and \approach one. 
Each of the evaluated method share the same Faster R-CNN architecture with a ResNet-50 backbone initialized with ImageNet pretrained weights \cite{deng2009imagenet}. We train the network using SGD optimizer with batch size equal to $4$, learning rate set to $10^{-3}$ and decreased after $75\%$ of iterations by a factor of $0.1$. The weight decay is set equal to $10^{-4}$ and SGD momentum to $0.9$. The number of training iterations is set to $40k$ for both the $1^{st}$ and the $2^{nd}$ training step, while it is decreased down to $10k$ in the $3^{rd}$ and $4^{th}$ step. 
% In the first stage of the training procedure, the NMS threshold for the RPN is set to $0.7$ and the number of top-$k$ scoring proposals to keep before applying NMS is set to 12.000 during training and 6.000 during test; after applying NMS it is set to 2.000 during training and 1.000 during test. The number of regions per image used to train the RPN is set to 256.
We resize images to $800 \times 1333$ in both training and testing, while during training we also performed random crop and random horizontal and vertical flip operations.

\begin{table*}[]
\caption{Evaluation of out-of-distribution detection methods adopting standard and \approach\ (ours) training strategies on Pascal VOC \cite{Everingham2009ThePV} and MS-COCO \cite{lin2014microsoft} datasets. Results on $\known$ are evaluated through mAP and $WI_{no\_rej}$ \cite{dhamija2020overlooked} metrics, while results on $\unknown$ are evaluated computing WI \cite{dhamija2020overlooked} and the recall, precision and F1 scores on unknown objects.
\vspace{10pt}}
\centering
\resizebox{\linewidth}{!}{
\begin{tabular}{l|l|c|cc|ccc}
\textbf{Training} & \textbf{Rejection} & \multicolumn{1}{c|}{\textbf{mAP\%$\uparrow$}} &
\multicolumn{1}{c}{\textbf{WI$_{no\_rej}\downarrow$}} &
\multicolumn{1}{c|}{\textbf{WI$\downarrow$}} & \multicolumn{1}{c}{\textbf{U$_{Recall}\uparrow$}} & \multicolumn{1}{c}{\textbf{U$_{Precision}\uparrow$}} & \multicolumn{1}{c}{\textbf{U$_{F1}\uparrow$}}  \\ \hline
Standard & Without Rejection \cite{ren2015faster} & 67.29 & 1.63 & 1.63 & 0.00 & 0.00 & 0.00  \\
 & MSP \cite{hendrycks2016baseline} & 67.36 & 1.58 & -17.52 & 2.46 & 4.20 & 3.10   \\
 & Energy \cite{liu2020energy} & 51.01 & 0.78 & \textbf{-30.39}  & 2.40 & 0.41 & 0.70 \\
 & ODIN \cite{liang2018enhancing} & 67.22 & 1.58 & -20.43 & 3.81 & 1.38 & 2.02  \\ \hline
\textbf{\approach} & Without Rejection  \cite{ren2015faster} & \textbf{67.75} & 1.50 & 1.50 & 0.00 & 0.00 & 0.00 \\
 & MSP \cite{hendrycks2016baseline} & 67.74 & 1.50 & -19.22 & 3.22 & \textbf{4.57} & \textbf{3.78} \\
 & Energy \cite{liu2020energy} & 51.75 & \textbf{0.68} & -29.82 & 2.34 & 0.30 & 0.53 \\
 & ODIN \cite{liang2018enhancing} & 67.63 & 1.49 & -21.56 & 4.85 & 1.62 & 2.43  \\ 
 & \neuron & \textbf{67.75} & 1.48 & -21.91 & \textbf{5.19} & 2.83 & 3.67 \\ \hline
\end{tabular}
}
\vspace{5pt}
\label{tab:results}
\end{table*}

\myparagraph{Metrics.} 
To assess the impact of the unknown objects on the performance of standard object detection models, \cite{dhamija2020overlooked} introduced the metric called \textit{Wilderness Impact (WI)}, defined as: 
% \small{
\begin{align} \label{WI-eq}
\begin{split}
        WI &= \frac{\text{Precision Closed-Set}}{\text{Precision Open-Set}} - 1 = \\ 
        &= \frac{TP_c}{TP_c + FP_c} \cdot \frac{TP_c + FP_c + TP_o + FP_o}{TP_c + TP_o} - 1, 
\end{split}
\end{align}
where $TP_c, FP_c$, indicate the true positives and false positives of known classes, while $TP_o, FP_o$ the true and false positive on unkowns.
However, in 
% where $P_{CS}$ and $P_{OS}$ are the precision on the closed and open set respectively.
\cite{dhamija2020overlooked} the metric was simplified since it did not considered the rejection option, \ie the models cannot predict the unknown class, thus $TP_o = 0$. They considered the following metric:
% \begin{equation} \label{WI-simplified}
%         WI_{simplified} = \frac{TP_c + FP_c + FP_o}{TP_c + FP_c} - 1 = \frac{FP_o}{TP_c + FP_c}
% \end{equation}
\begin{equation} \label{WI-simplified}
        WI_{no\_rej} = \frac{FP_o}{TP_c + FP_c}.
\end{equation}
We note that $WI_{no\_rej}$ does not consider the detection performance on unknown classes, but it considers only the performance on the known classes. In particular, classifying all the unknown as background objects would get the optimal score since $FP_o=0$. 
Despite the second metric has been used by recent works \cite{joseph2021towards, dhamija2020overlooked}, we consider more suited to our task the $WI$ metric since it took also into account the true positive on unknown objects. 

Moreover, since in our work we are explicitly interested 
% In our work, however, we are interested 
in evaluating the models on unknown objects, we report the recall, precision and F1 metrics considering only the unknown class.
%, \ie we consider $TP_o \neq 0$. Providing to the models the capability of rejection, $WI$ formula is not sufficient to capture unknown performances. Thus, for a complete evaluation, we compute recall, precision and F1 metrics on the unknown class.
% \myparagraph{U-Recall, U-Precision, U$_{F1}$.}
% In order to measure how the model is at the same time accurate in detecting unknowns (i.e. \textit{precision}) and good at finding unknown objects (i.e. \textit{recall}), 
We define them as:
\begin{equation} \label{u-recall}
    U_{Recall} = \frac{TP_o}{TP_o + FN_o},
\end{equation}
\begin{equation} \label{u-precision}
   U_{Precision} = \frac{TP_o}{TP_o + FP_o}.
\end{equation}
The more the model tends to predict each sample as belonging to the unknown category, the higher the recall will be, but the lower the precision will be as the number of false positive tends to increase. For this reason, we introduce the \textit{$U_{F1}$} metric which summarizes both $U_{Recall}$ and $U_{Precision}$. $U_{F1}$ is maximized if and only if both $U_{Recall}$ and $U_{Precision}$ are maximized. It is defined as:
\begin{equation}
    U_{F1} = 2 \cdot \frac{U_{Recall} \cdot U_{Precision}}{U_{Recall} + U_{Precision}}.
\end{equation}

% \textbf{UD-\textit{Recall}, UD-\textit{Precision}, UD$_{F1}$.} In object detection we can define these two types of error (\ref{fig:udr-udp}):

% \begin{enumerate}
%     \item \textbf{Missing detection:} The object was not located, and consequently not labeled as one of the classes.
    
%     \item \textbf{Misclassification:} The object was correctly located, but the predicted label is different from the ground-truth.
% \end{enumerate}

% Typical metrics such as \textit{recall} and \textit{precision} equally treats these errors, although the second type of error is intituitively less serious than a missing detection. For this reason, \cite{revisiting-owod} introduces two metrics: $UD-Recall$ and $UD-Precision$, that are defined as:

% \begin{equation} \label{ud-recall}
%     UD-Recall = \frac{TP_o + FN^*_o}{TP_o + FN_o}
% \end{equation}
% \begin{equation} \label{ud-precision}
%     UD-Precision = \frac{TP_o}{TP_o + FN^*_o}
% \end{equation}

% where $FN^*_o$ indicates the number of ground-truth boxes recalled by misclassified predicted bounding boxes. For the same reason explained above, we introduce $UD$_{F1}$$ that is:

% \begin{equation}
%     UD$_{F1}$ = 2 \cdot \frac{UD-Recall \cdot UD-Precision}{UD-Recall + UD-Precision}
% \end{equation}

% \myparagraph{mAP}. 
Finally, in addition to the open-set metrics, we also report the \textit{mAP} to evaluate the model performances on $\mathcal{Y}$ classes, defined as:
\begin{equation} \label{map-equation}
    mAP = \frac{1}{|\mathcal{Y}|} \displaystyle\sum_{c \in \mathcal{Y}}{AP_c},
\end{equation}
where $AP_c$ is the average precision for the class $c$ computed at different recall levels.

\subsection{Unknown detection results}
\cref{tab:results} reports the comparison among the standard Faster R-CNN training and \approach, considering multiple unknown detection strategies.
As the table shows, our approach improves detection of both known and unknown objects. Starting from the former, standard Faster R-CNN architecture with no rejection capability benefits from our approach, achieving up to to 67.75\% mAP. This behaviour is also confirmed by the WI improvement from 1.63 to 1.5 (the lower the better), indicating that \approach\ distinguishes better known and unknown objects. The same behavior on the known classes is evident when employing a rejection strategy: considering the WI metric, MSP improves from -17.52 to -19.22 and ODIN from -20.43 to -21.56, while Energy obtain similar results (-30.39 vs -29.82). We note that, while the Energy approach is the best on the WI and WI$_{no\_rej}$ metric, it obtains very low mAP. We ascribe this behavior to the high unknown score that Energy assigns to samples, leading to the rejection of most of the known objects. Moreover, the contrast among the WI and mAP metrics reveals that WI is not well-suited to evaluate methods on the open-set task, since it does not consider the overall model performance but only the ratio among closed- and open-set precision.

While improving the results on known classes is important, our goal is mainly detect unknown objects. Considering the $U$-$F1$ score, we note that \approach\ increases the results of MSP from $3.10\%$ to $3.78\%$ and from $2.02\%$ to $2.43\%$ for ODIN \cite{liang2018enhancing}. This remarks the impressive ability of our training strategy, that improves the detection ability of the model without introducing additional costs, pseudo-labeling unknown objects in the background of the images.
We acknowledge that our training procedure slightly hamper the Energy performance. However, we remark that Energy already assigns a very high score for the objects, as demonstrated by the very low $U_{Precision}$ achieved by it. We also note that only Energy shows this behavior, while both MSP and ODIN, when using \approach, improve the results on both $U_{Precision}$ and $U_{Recall}$ metrics.
%Regarding the , our approach enhances also out-of-distribution strategies, increasing the $U$-$F1$ score from $3.10\%$ to $3.78\%$ for MSP \cite{hendrycks2016baseline} and from $2.02\%$ to $2.43\%$ for ODIN \cite{liang2018enhancing}, while Energy \cite{liu2020energy} performance slightly decreases.
%\dario{tagliare sto pippone e magari metterlo nella rejection di energy} The reasons of its poor results with both the training procedures is due to implementation constraints. Indeed, in order to compute Non-Maximum-Suppression (NMS) \cite{girshick2015fast, ren2015faster} on unknown object proposals, we have been forced to normalize the energy of each proposal to have it between $0$ and $1$, with a consequent loss of information. It also confirms the inapplicability of WI measure to evaluate unknown predictions, as for Energy it is almost zero, but this is a misleading result as the mAP is, instead, very low. The reason is that Energy achieves very similar precision in both closed- and open-set evaluations and thus minimizes WI, but having those precision values very low, also the mAP decreases a lot. Moreover, our \approach approach allows also the direct prediction of the unknown class, reaching performance in accordance with standard rejection strategies, achieving up to the highest $U$-$Recall$ of 5.19\%. 
Finally, comparing the simple direct prediction approach with the others, we see that it achieves comparable or even better performance. In particular, it achieves the best performance on $U_{Recall}$ without suffering performance drop, as indicated by the mAP performance. However, the $U_{Precision}$ is lower than other approaches (-1.74\% \wrt MSP). % In the following, we further investigate the \neuron\ approach.

\begin{table}[t]
\caption{Ablation study of the direct approach rejection strategy when adopting \approach. We compute mAP and U$_{F1}$ metrics without rejection, using the direct prediction of the final classifier, exploiting the $\tau_{obj}$ for unknown predictions and combining the two strategies.}

\centering
\resizebox{\linewidth}{!}{
\begin{tabular}{ccc|cc}
\textbf{No rejection} & \textbf{$\hat{y} = \unknown$} & \textbf{$\tau_{obj}$} & \textbf{mAP$\uparrow$} & \textbf{U$_{F1} \uparrow$} \\ \hline % \textbf{WI_{no\_rej}$\downarrow$} & 
\ding{51} &  &  & 67.75 & 0.00 \\
 & \ding{51} &  & 67.73 & 3.46 \\
 & & \ding{51}  & 67.72 & 3.41 \\ \hline
 & \ding{51} & \ding{51} & \textbf{67.75} &\textbf{ 3.67} \\ \hline
\end{tabular} \label{tab:ablation_unk}
}
\end{table}
\subsection{Ablation studies} \label{sec:ablation}
\myparagraph{Direct prediction rejection strategy}.
In Table \ref{tab:ablation_unk} we report a detailed analysis on the direct prediction rejection strategy available when using our \approach\ approach. It is composed by two key components: the additional unknown class on the final classification layer, able to predict the unknown class for each RoI; and the objectness threshold $\tau_{obj}$. As we can see from the second row, the additional unknown class allows to maintain the same $67.75\%$ mAP performance over the known classes of the standard Faster R-CNN with no rejection capability; while it also achieves 3.46\% of $U$-$F1$. 
In the third row, we reported instead the results achieved by directly applying $\tau_{obj}$ to background RoIs. In particular, for each RoI predicted as background with the highest probability among all classes, if the $\detector_{RPN}$ emits an objectness score higher than $\tau_{obj}$ it is then considered unknown. Combining the two strategies together allows to achieve the best results, as shown in the last row of the table. The overall rejection procedure maintains the highest mAP on known classes, while also increases the $U$-$F1$ up to $3.67\%$ and decreases the WI down to $1.48$.

 \begin{figure*}[t]
    \centering
    \includegraphics[width=\linewidth]{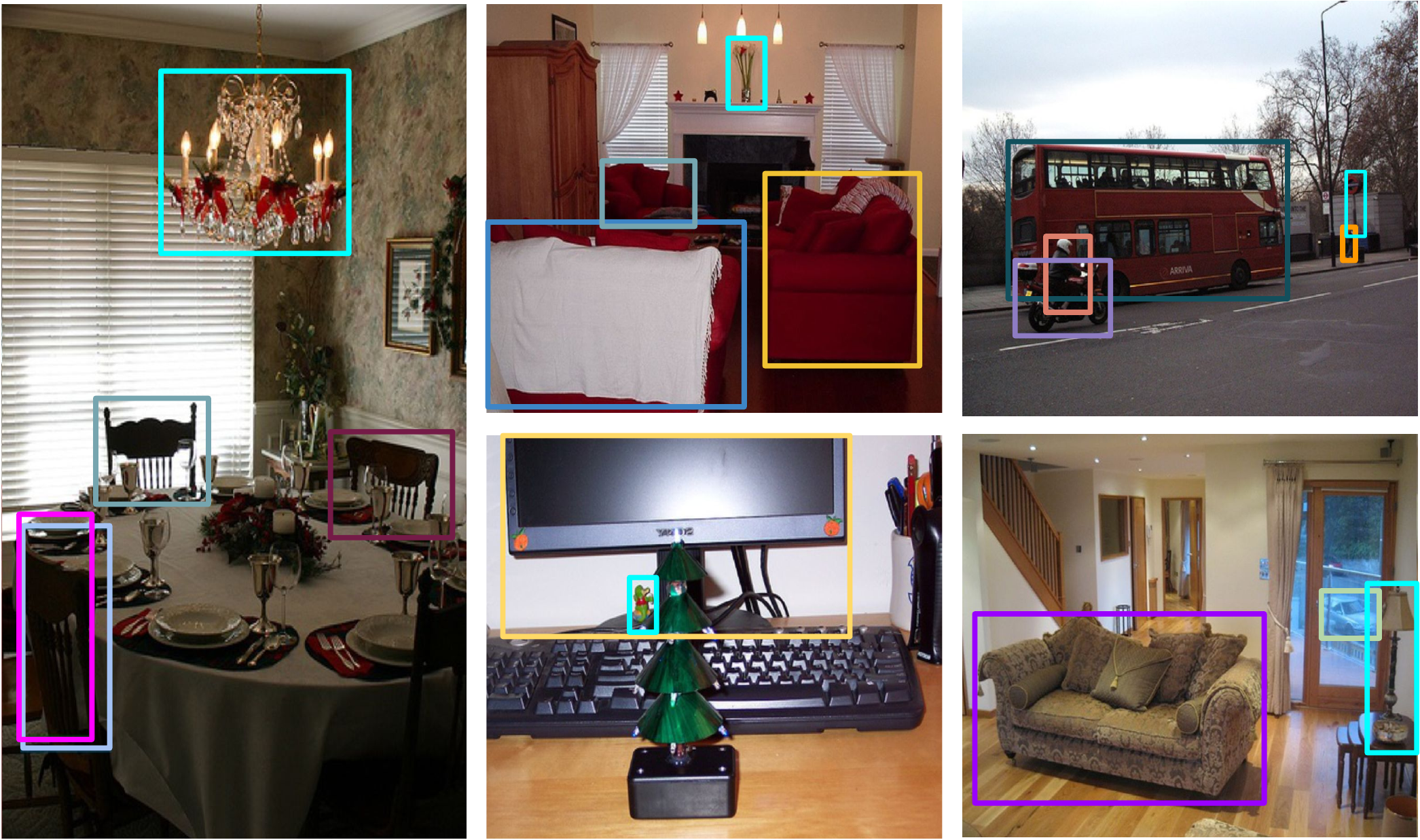}
   \caption{\textbf{Qualitative} results of $\detector_{RPN}$ detections on Pascal VOC \cite{Everingham2009ThePV}. The cyan boxes indicate a detection on unknown objects, while the other ones indicate a detection on known classes. Best viewed in color.}%   
   \label{fig:qualitative}
 \end{figure*}

% \comment{
\myparagraph{RPN unknown detection.} Although $\detector_{RPN}$ is a class-agnostic detector by design \cite{ren2015faster, girshick2015fast}, an open question is whether it is able to recognize even objects on which it has not been explicitly trained on, as it is essential for the unknown pseudo-labeling procedure. To this end, we formulated the $AVG_{obj}$ metric as a quantitative measure of the $\detector_{RPN}$ ability to identify objects within an image in both closed- and open-set scenarios. It is formulated as follows:
\begin{equation} \label{mean-rpn}
  AVG_{obj} = \frac{1}{|\mathcal{P}_{fg}|}\displaystyle\sum_{p \in \mathcal{P}_{fg}} f_{fg}(p)
\end{equation}
where $\mathcal{P}_{fg}$ is the set of ground truth foreground proposals and $f_{fp}(p)$ is the probability
% given by the $\detector_{RPN}$ 
that the proposal $p$ is actually considered foreground.

% Since it is calculated when the candidate proposal is labeled as foreground, it gives a quantitative measure of the ability of the RPN to identify objects within an image. The higher $HM_{obj}$ is, the more the $\detector_{RPN}$ is able to separate the background from the foreground. 
We report in \cref{tab:mean-rpn} the evaluation of standard training procedure and \approach\  under the $AVG_{obj}$ computed on both known and unknown objects. Our approach achieves up to $0.99$ on known and $0.98$ on unknown (being 1 the upper bound), surpassing the standard procedure in both the evaluations. Achieving comparable performance in both closed- and open-set scenarios proves our intuition that $\detector_{RPN}$ is able to precisely detect objects despite their belonging to the training distribution or not. It is worth noting that \approach\ increases the confidence on considering proposals as foreground ones on the known classes, as shown in the comparison between the first row and the third one.

% \begin{table}[t]
% \caption{Caption}
% \centering
% \resizebox{\linewidth}{!}{
% \begin{tabular}{c|cc|c|c|c}
% $\detector_{RPN}$ & Standard & \approach & $\known$ & $\unknown$ & \multicolumn{1}{l}{$HS_{obj}\uparrow$} \\ \hline
% Step 1 & \ding{51} &  & \ding{51} &  & 87.68 \\
% \textbf{} & \ding{51} &  &  & \ding{51} & 77.12 \\ \hline
% Step 3 & \ding{51} &  & \ding{51} &  & 86.44 \\
% \textbf{} & \ding{51} &  &  & \ding{51} & 77.55 \\
% Step 3 &  & \ding{51} & \ding{51} &  & \textbf{89.27} \\
% \textbf{} &  & \ding{51} &  & \ding{51} & \textbf{82.81} \\ \hline
% \end{tabular}
% }\label{tab:harmonic-mean-rpn}
% \end{table}

\begin{table}[t]
\caption{Ablation study of $\detector_{RPN}$ ability to identify objects in closed- and open-set scenarios.}
\centering
% \resizebox{0.75\linewidth}{!}{
\begin{tabular}{cc|cc|c}
Standard & \approach & known & unknown & $AVG_{obj}$ \\ \hline
\ding{51} &  & \ding{51} &  & 0.98 \\
\ding{51} &  &  & \ding{51} & 0.97 \\
 & \ding{51} & \ding{51} &  & \textbf{0.99} \\
 & \ding{51} &  & \ding{51} & 0.98 \\ \hline
\end{tabular}
\label{tab:mean-rpn}
\end{table}

% \begin{table}[t]
% \centering
% \resizebox{\linewidth}{!}{
% \begin{tabular}{l|ccc}
% Method & \multicolumn{1}{c}{AUPR $\uparrow$} & \multicolumn{1}{c}{AUROC $\uparrow$} & \multicolumn{1}{c}{FPR95 $\downarrow$} \\ \hline
% AE \cite{baur2018deep} & 2.2 & 66.1 & 91.7 \\
% Dropout \cite{gal2016dropout} & 7.5 & 69.9 & 79.4 \\
% MSP \cite{hendrycks2016baseline} & 6.6 & 87.7 & 33.7 \\
% MSP + CRF \cite{hendrycks2019benchmark} & 6.5 & 88.1 & 29.9 \\
% SynthCP \cite{xia2020synthesize} & \textbf{9.3} & 88.5 & 28.4 \\ \hline
% \textbf{\ours} & 8.8 & \textbf{91.1} & \textbf{23.2}
% \end{tabular}}
% \vspace{-8pt}
% \caption{Results on StreetHazards dataset \cite{hendrycks2019benchmark} according to AUPR, AUROC and FPR95 metrics. \cite{hendrycks2019benchmark}.
% \vspace{-10pt}
% }
% \label{tab:streethazards}
% \end{table}

% \begin{figure*}[t]
%     \centering
%     \includegraphics[width=0.94\linewidth]{images/qualitative2.png}
%     \vspace{-5pt}
%      \caption{\textbf{Qualitative comparison} between the use of probabilities (MSP) and our direct scores (PAnS) for segmenting anomalies on StreetHazards \cite{hendrycks2019benchmark}. White indicates an high score for the anomaly, while the blue indicates a low score. In the semantic labels the anomaly are represented in cyan.}
%     \vspace{-5pt}
%     \label{fig:qualitative}
%  \end{figure*}

\section{Conclusions}
\label{sec:conclusions}
In this work, we proposed a novel training strategy, called \approach, to improve open-set object detection performance. \approach\ relies on the assumption that, during training, the images contain multiple non-annotated objects. Instead of requiring an explicit annotation for them, it automatically detects and pseudo-labels them, exploiting the four-steps Faster R-CNN training procedure. In particular, in the first step, it trains the class-agnostic RPN to detect objects using the ground truth annotations. Then, it pseudo-labels as unkown all the objects in the dataset with a high objectness score that do not match a ground truth annotation. The pseudo-labels are then used as pseudo ground-truths to train the classification head. In the third and fourth training steps, the knowledge on the unknowns is further consolidated, obtaining the final model.

We demonstrate that \approach\ is able to directly detect the unknown classes and it also improves the performance of previous training strategies with no additional costs on the Pascal VOC and MS-COCO datasets.
Indeed, the unknown detection performance is still far from a system ready to operate in the wild and we hope that our work establishes a new baseline to push forward the state of the art in this  research field.

%%%%%%%%% REFERENCES
{\small
\bibliographystyle{ieee_fullname}
\bibliography{egbib}

\begin{thebibliography}{10}\itemsep=-1pt

\bibitem{bendale2015towards}
Abhijit Bendale and Terrance Boult.
\newblock Towards open world recognition.
\newblock In {\em CVPR-15}.

\bibitem{cao2019hierarchical}
Jiale Cao, Yanwei Pang, Jungong Han, and Xuelong Li.
\newblock Hierarchical shot detector.
\newblock In {\em Proceedings of the IEEE/CVF International Conference on
  Computer Vision}, pages 9705--9714, 2019.

\bibitem{corbiere2019addressing}
Charles Corbi{\`e}re, Nicolas Thome, Avner Bar-Hen, Matthieu Cord, and Patrick
  P{\'e}rez.
\newblock Addressing failure prediction by learning model confidence.
\newblock In {\em Adv. Neural Inform. Process. Syst.}, pages 2902--2913, 2019.

\bibitem{deng2009imagenet}
Jia Deng, Wei Dong, Richard Socher, Li-Jia Li, Kai Li, and Li Fei-Fei.
\newblock Imagenet: A large-scale hierarchical image database.
\newblock In {\em 2009 IEEE conference on computer vision and pattern
  recognition}, pages 248--255. Ieee, 2009.

\bibitem{dhamija2020overlooked}
Akshay Dhamija, Manuel Gunther, Jonathan Ventura, and Terrance Boult.
\newblock The overlooked elephant of object detection: Open set.
\newblock In {\em Proceedings of the IEEE/CVF Winter Conference on Applications
  of Computer Vision}, pages 1021--1030, 2020.

\bibitem{dhamija2018reducing}
Akshay~Raj Dhamija, Manuel G{\"u}nther, and Terrance Boult.
\newblock Reducing network agnostophobia.
\newblock {\em Advances in Neural Information Processing Systems}, 31, 2018.

\bibitem{Everingham2009ThePV}
Mark Everingham, Luc~Van Gool, Christopher K.~I. Williams, John~M. Winn, and
  Andrew Zisserman.
\newblock The pascal visual object classes (voc) challenge.
\newblock {\em International Journal of Computer Vision}, 88:303--338, 2009.

\bibitem{gal2016dropout}
Yarin Gal and Zoubin Ghahramani.
\newblock Dropout as a bayesian approximation: Representing model uncertainty
  in deep learning.
\newblock In {\em ICML}, pages 1050--1059, 2016.

\bibitem{girshick2015fast}
Ross Girshick.
\newblock Fast r-cnn.
\newblock In {\em Proceedings of the IEEE international conference on computer
  vision}, pages 1440--1448, 2015.

\bibitem{girshick2014rich}
Ross Girshick, Jeff Donahue, Trevor Darrell, and Jitendra Malik.
\newblock Rich feature hierarchies for accurate object detection and semantic
  segmentation.
\newblock In {\em Proceedings of the IEEE conference on computer vision and
  pattern recognition}, pages 580--587, 2014.

\bibitem{gupta2021ow}
Akshita Gupta, Sanath Narayan, KJ Joseph, Salman Khan, Fahad~Shahbaz Khan, and
  Mubarak Shah.
\newblock Ow-detr: Open-world detection transformer.
\newblock {\em arXiv preprint arXiv:2112.01513}, 2021.

\bibitem{he2017mask}
Kaiming He, Georgia Gkioxari, Piotr Doll{\'a}r, and Ross Girshick.
\newblock Mask r-cnn.
\newblock In {\em Proceedings of the IEEE international conference on computer
  vision}, pages 2961--2969, 2017.

\bibitem{hendrycks2016baseline}
Dan Hendrycks and Kevin Gimpel.
\newblock A baseline for detecting misclassified and out-of-distribution
  examples in neural networks.
\newblock {\em arXiv preprint arXiv:1610.02136}, 2016.

\bibitem{hendrycks2018deep}
Dan Hendrycks, Mantas Mazeika, and Thomas Dietterich.
\newblock Deep anomaly detection with outlier exposure.
\newblock {\em arXiv preprint arXiv:1812.04606}, 2018.

\bibitem{hsu2020generalized}
Yen-Chang Hsu, Yilin Shen, Hongxia Jin, and Zsolt Kira.
\newblock Generalized odin: Detecting out-of-distribution image without
  learning from out-of-distribution data.
\newblock In {\em IEEE Conf. Comput. Vis. Pattern Recog.}, pages 10951--10960,
  2020.

\bibitem{joseph2021towards}
KJ Joseph, Salman Khan, Fahad~Shahbaz Khan, and Vineeth~N Balasubramanian.
\newblock Towards open world object detection.
\newblock In {\em Proceedings of the IEEE/CVF Conference on Computer Vision and
  Pattern Recognition}, pages 5830--5840, 2021.

\bibitem{kendall2017uncertainties}
Alex Kendall and Yarin Gal.
\newblock What uncertainties do we need in bayesian deep learning for computer
  vision?
\newblock In {\em Adv. Neural Inform. Process. Syst.}, pages 5574--5584, 2017.

\bibitem{lee2018simple}
Kimin Lee, Kibok Lee, Honglak Lee, and Jinwoo Shin.
\newblock A simple unified framework for detecting out-of-distribution samples
  and adversarial attacks.
\newblock In {\em Adv. Neural Inform. Process. Syst.}, pages 7167--7177, 2018.

\bibitem{li2020background}
Yi Li and Nuno Vasconcelos.
\newblock Background data resampling for outlier-aware classification.
\newblock In {\em Proceedings of the IEEE/CVF Conference on Computer Vision and
  Pattern Recognition}, pages 13218--13227, 2020.

\bibitem{liang2018enhancing}
Shiyu Liang, Yixuan Li, and R Srikant.
\newblock Enhancing the reliability of out-of-distribution image detection in
  neural networks.
\newblock In {\em International Conference on Learning Representations}, 2018.

\bibitem{lin2017feature}
Tsung-Yi Lin, Piotr Doll{\'a}r, Ross Girshick, Kaiming He, Bharath Hariharan,
  and Serge Belongie.
\newblock Feature pyramid networks for object detection.
\newblock In {\em Proceedings of the IEEE conference on computer vision and
  pattern recognition}, pages 2117--2125, 2017.

\bibitem{lin2017focal}
Tsung-Yi Lin, Priya Goyal, Ross Girshick, Kaiming He, and Piotr Doll{\'a}r.
\newblock Focal loss for dense object detection.
\newblock In {\em Proceedings of the IEEE international conference on computer
  vision}, pages 2980--2988, 2017.

\bibitem{lin2014microsoft}
Tsung-Yi Lin, Michael Maire, Serge Belongie, James Hays, Pietro Perona, Deva
  Ramanan, Piotr Doll{\'a}r, and C~Lawrence Zitnick.
\newblock Microsoft coco: Common objects in context.
\newblock In {\em European conference on computer vision}, pages 740--755.
  Springer, 2014.

\bibitem{liu2016ssd}
Wei Liu, Dragomir Anguelov, Dumitru Erhan, Christian Szegedy, Scott Reed,
  Cheng-Yang Fu, and Alexander~C Berg.
\newblock Ssd: Single shot multibox detector.
\newblock In {\em European conference on computer vision}, pages 21--37.
  Springer, 2016.

\bibitem{liu2020energy}
Weitang Liu, Xiaoyun Wang, John Owens, and Yixuan Li.
\newblock Energy-based out-of-distribution detection.
\newblock {\em Advances in Neural Information Processing Systems},
  33:21464--21475, 2020.

\bibitem{miller2018dropout}
Dimity Miller, Lachlan Nicholson, Feras Dayoub, and Niko S{\"u}nderhauf.
\newblock Dropout sampling for robust object detection in open-set conditions.
\newblock In {\em 2018 IEEE International Conference on Robotics and Automation
  (ICRA)}, pages 3243--3249. IEEE, 2018.

\bibitem{peng2020faster}
Can Peng, Kun Zhao, and Brian~C Lovell.
\newblock Faster ilod: Incremental learning for object detectors based on
  faster rcnn.
\newblock {\em Pattern Recognition Letters}, 140:109--115, 2020.

\bibitem{perera2020generative}
Pramuditha Perera, Vlad~I Morariu, Rajiv Jain, Varun Manjunatha, Curtis
  Wigington, Vicente Ordonez, and Vishal~M Patel.
\newblock Generative-discriminative feature representations for open-set
  recognition.
\newblock In {\em Proceedings of the IEEE/CVF Conference on Computer Vision and
  Pattern Recognition}, pages 11814--11823, 2020.

\bibitem{redmon2016you}
Joseph Redmon, Santosh Divvala, Ross Girshick, and Ali Farhadi.
\newblock You only look once: Unified, real-time object detection.
\newblock In {\em Proceedings of the IEEE conference on computer vision and
  pattern recognition}, pages 779--788, 2016.

\bibitem{redmon2017yolo9000}
Joseph Redmon and Ali Farhadi.
\newblock Yolo9000: better, faster, stronger.
\newblock In {\em Proceedings of the IEEE conference on computer vision and
  pattern recognition}, pages 7263--7271, 2017.

\bibitem{ren2015faster}
Shaoqing Ren, Kaiming He, Ross Girshick, and Jian Sun.
\newblock Faster r-cnn: Towards real-time object detection with region proposal
  networks.
\newblock {\em Advances in neural information processing systems}, 28, 2015.

\bibitem{shmelkov2017incremental}
Konstantin Shmelkov, Cordelia Schmid, and Karteek Alahari.
\newblock Incremental learning of object detectors without catastrophic
  forgetting.
\newblock In {\em Proceedings of the IEEE international conference on computer
  vision}, pages 3400--3409, 2017.

\bibitem{vyas2018out}
Apoorv Vyas, Nataraj Jammalamadaka, Xia Zhu, Dipankar Das, Bharat Kaul, and
  Theodore~L Willke.
\newblock Out-of-distribution detection using an ensemble of self supervised
  leave-out classifiers.
\newblock In {\em Proceedings of the European Conference on Computer Vision
  (ECCV)}, pages 550--564, 2018.

\bibitem{wang2019learning}
Tiancai Wang, Rao~Muhammad Anwer, Hisham Cholakkal, Fahad~Shahbaz Khan, Yanwei
  Pang, and Ling Shao.
\newblock Learning rich features at high-speed for single-shot object
  detection.
\newblock In {\em Proceedings of the IEEE/CVF International Conference on
  Computer Vision}, pages 1971--1980, 2019.

\bibitem{zhou2021learning}
Da-Wei Zhou, Han-Jia Ye, and De-Chuan Zhan.
\newblock Learning placeholders for open-set recognition.
\newblock In {\em Proceedings of the IEEE/CVF Conference on Computer Vision and
  Pattern Recognition}, pages 4401--4410, 2021.

\end{thebibliography}
}

\end{document}